\pdfoutput=1

\documentclass[11pt]{article}

\usepackage{etoolbox}


\usepackage[final]{acl}

\usepackage{times}
\usepackage{latexsym}

\usepackage[T1]{fontenc}

\usepackage[utf8]{inputenc}

\usepackage{microtype}

\usepackage{inconsolata}
\usepackage{multirow}
\usepackage{multicol}  

\usepackage{graphicx}

\usepackage{booktabs}   
\usepackage{adjustbox}  

\title{Basic Reading Distillation}

\author{
  \textbf{Zhi Zhou\textsuperscript{1}}\thanks{Equal Contribution},
  \textbf{Sirui Miao\textsuperscript{1$\ast$}},
  \textbf{Xiangyu Duan\textsuperscript{1}\thanks{Corresponding Author}},
  \textbf{Hao Yang\textsuperscript{2}},
  \textbf{Min Zhang\textsuperscript{1}},
\\
  \textsuperscript{1}School of Computer Science and Technology, Soochow University, Suzhou, China
\\
  \textsuperscript{2}Huawei Translation Services Center, Beijing, China
\\
  \small{
    \textbf{\{zzhou2002,srmiao\}@stu.suda.edu.cn; \{xiangyuduan,minzhang\}@suda.edu.cn; yanghao30@huawei.com}
  }
}

\begin{document}
\maketitle
\begin{abstract}
Large language models (LLMs) have demonstrated remarkable abilities in various natural language processing areas, but they demand high computation resources which limits their deployment in real-world. Distillation is one technique to solve this problem through either knowledge distillation or task distillation. Both distillation approaches train small models to imitate specific features of LLMs, but they all neglect basic reading education for small models on generic texts that are \emph{unrelated} to downstream tasks. In this paper, we propose basic reading distillation (BRD) which educates a small model to imitate LLMs basic reading behaviors, such as named entity recognition, question raising and answering, on each sentence. After such basic education, we apply the small model on various tasks including language inference benchmarks and BIG-bench tasks. It shows that the small model can outperform or perform comparable to over 20x bigger LLMs. Analysis reveals that BRD effectively influences the probability distribution of the small model, and has orthogonality to either knowledge distillation or task distillation.

\end{abstract}

\section{Introduction}

Large language models (LLMs) exhibit consistent performance gains across various areas \cite{zhao2023survey,huang-chang-2023-towards,chang2023survey}. Nevertheless, their formidable size and high computational requirements impede their real-world applications. Distillation is one widespread approach to tackle this issue by distilling LLMs into smaller language models. It is divided into mainly two categories: knowledge distillation and task distillation. Both distillation approaches  adopt the teacher-student framework, in which the smaller language models act as the student models, and are trained to imitate specific features of LLMs, which act as the teacher models. Specifically, knowledge distillation \cite{hinton2015distilling} usually trains the student models to imitate implicit features inside the teacher models, while task distillation \cite{chen2020big} usually trains the student models to imitate explicit behaviors of the teacher models.

\begin{figure}[t!]
    \centering

    \includegraphics[width=0.5\textwidth]{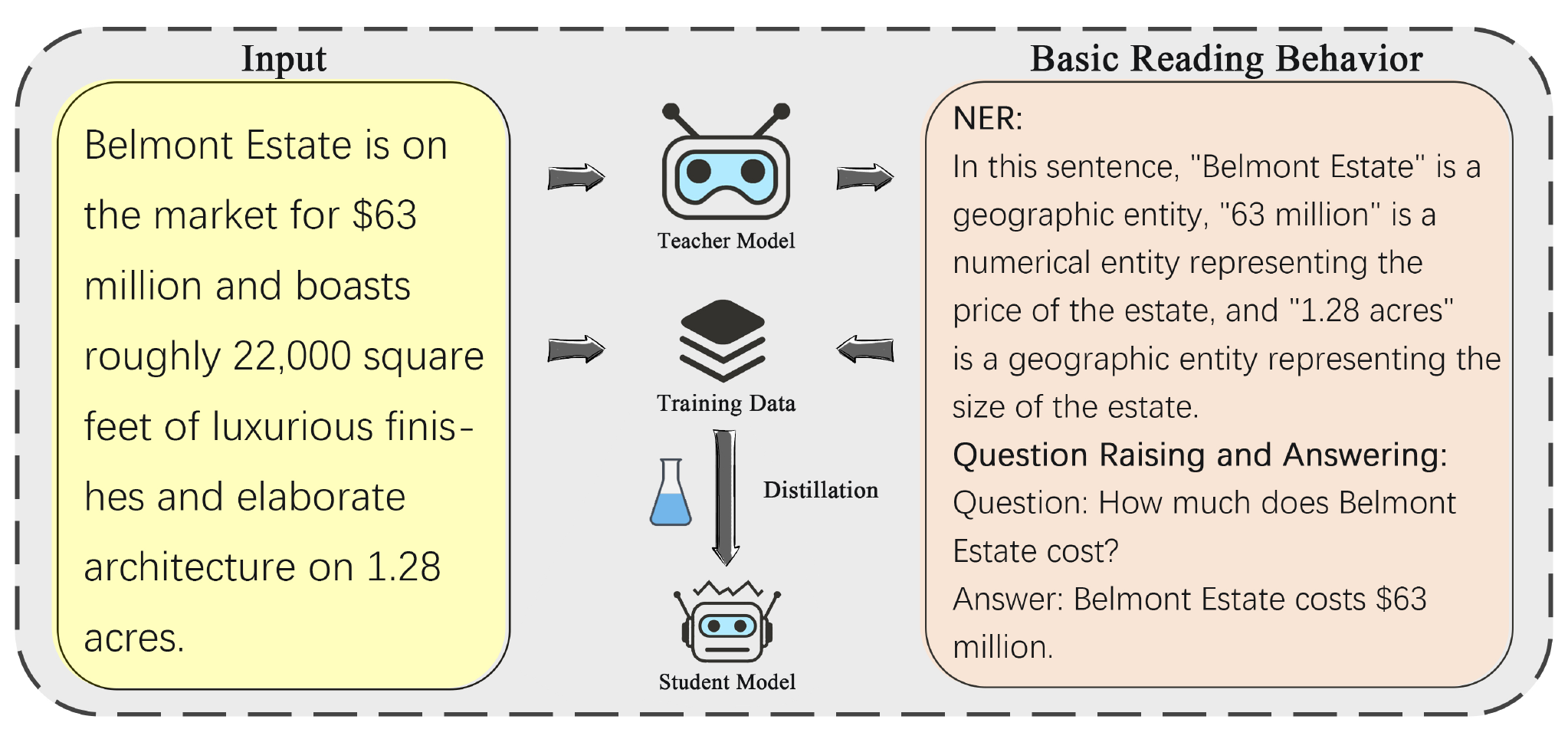}

    \caption{The illustration of BRD process.}
    \label{fig:framework}
\end{figure}

Different to both distillation approaches, we propose basic reading distillation (BRD) that teaches a student model basic reading abilities such as named entity recognition, question raising, and question answering, on general sentences. It simulates human reading education via interactions including raising questions about parts of a sentence, answering the questions, extracting important information such as named entities. Such basic reading education on every sentence is important before application on downstream tasks, while is always neglected in both knowledge distillation and task distillation.

Our motivation is that LLMs should be educated beyond just trained for next token prediction. The education brings LLMs  basic reading ability in text comprehension. It is like that LLMs experience the high school/college education on reading, then is well prepared for the future test, while traditional LLMs training just consumes tokens, then let LLMs directly participate in the future test. Regarding the distillation, we think that one well educated student is more effective than the one without reading education.

Furthermore, the benefits of BRD are two-fold: Firstly, beyond only using texts for training next token prediction, BRD educates the student model to deeply understand the texts via interactions. All available data such as web mined corpora can be extended to be magnitudes larger by BRD, breaking the data scale and diversity limitation criticized by \citet{gudibande2023false} in the task distillation. Secondly, BRD also avoids the implicit nature of knowledge distillation which imitates latent features such as logits \cite{hinton2015distilling}, hidden layers \cite{jiao-etal-2020-tinybert}, and attention maps \cite{li-etal-2020-bert,wang-etal-2021-minilmv2}. Such implicit nature leads to the deficiency of learning interpretability, while BRD demonstrates explicit reading behaviors that are easy to interpret. 



Figure \ref{fig:framework} illustrates the process of BRD. It starts by prompting LLMs to generate basic reading behaviors on general sentences, then proceeds with training the student model to imitate these behaviors. Experiments on various NLP tasks, including language inference benchmarks and Google Big-bench tasks, show that although the student model is trained on the general data that is irrelevant to the downstream tasks, it can inherit teacher model abilities, leading to excellent downstream performances better than or comparable to those of larger models. Furthermore, after this basic education of the student model on general sentences, we fine-tune the student model for downstream tasks, and find that the basic reading education leads to further improvement on downstream tasks, achieving on par or better performances when compared to the over 20x bigger teacher model. To analyze the effect of BRD, we compute the cross entropy between the student model and the teacher model, and find that the student model distribution approaches closer to the teacher model distribution after BRD, leading to better performances than non-educated ones. In summary, the main contributions are: 

\begin{itemize}

\item We propose BRD that educates the student model to imitate basic reading behaviors of the teacher model.

\item Experiments show that the student model exhibits excellent abilities distilled from the teacher model on various downstream tasks, achieving on par or even better performances against the teacher model.

\item The analysis reveals that BRD can drive the student model distribution closer to the teacher model distribution, resulting in significant performance improvements.

\end{itemize}

\section{Related Works}

There are mainly two streams of distillation approaches: knowledge distillation and task distillation. Knowledge distillation focuses on teaching implicit features inside the teacher model, while task distillation focuses on teaching explicit behaviors of the teacher model on downstream tasks. In addition, we introduce intrinsic task pre-training that focuses on intrinsic task data derived from the training plain texts. 

\subsection{Knowledge Distillation}

The field is pioneered by \citet{Bucila2006ModelC,hinton2015distilling}, followed by works using various types of internal information from the teacher model, including attention maps \cite{li-etal-2020-bert,wang-etal-2021-minilmv2}, output logits \cite{liu-etal-2020-fastbert}, hidden layers \cite{jiao-etal-2020-tinybert}. In the era of LLMs, GKD uses advanced memory optimization methods to address the memory constraint problem in distilling LLMs \cite{tan-etal-2023-gkd}, MiniLLM uses reverse KL divergence to prevent the student model from overestimating the void regions of the teacher distribution \cite{Gu2023KnowledgeDO}. \citet{agarwal2024onpolicy} use on-policy distillation that trains the student model on its self-generated mistakes. In the case that internal information of LLMs is not accessible and only decisions of LLMs are available, \citet{zhou-etal-2023-bridging} estimate logits from the decision distributions to train the student model.

\subsection{Task Distillation}

The task predictions or reasoning rationales made by the teacher model are used to train the student model in task distillation. Despite the noisy predictions of the teacher model, the student model achieves good imitation effects in performing the tasks \cite{chen2020big,wang-etal-2021-want-reduce,iliopoulos2022weighted,agrawal2023qameleon}. Besides the task predictions, rationales for the answers generated by the teacher model show efficiency in training the student model with less data \cite{hsieh-etal-2023-distilling,wang2023pinto,ho-etal-2023-large,magister-etal-2023-teaching}. Task distillation is closely related to model imitation researches \cite{orekondy19knockoff,wallace-etal-2020-imitation}, which collect API outputs of a a proprietary LM for some tasks, then use the outputs to fine-tune an open-source LM. \citet{gudibande2023false} criticize the data scale and the limited diversity in model imitation. \citet{mukherjee2023orca} address this criticism by using explanation tuning, more task data, and instructions. In comparison, BRD can perform on every sentence, leading to unlimited data resource that breaks the limitation on data scale and diversity.

In summary, task distillation focuses on the data of specific downstream tasks, while our BRD mainly focuses on general sentences unrelated to any specific downstream tasks, and the basic reading behaviors in BRD are basic education resource not aiming at any specific applications.

\subsection{Intrinsic Task Pre-training}

Different to task distillation that utilizes downstream task data, intrinsic task pre-training uses general training set to synthesize task data. PICL is a framework focusing on intrinsic tasks \cite{gu-etal-2023-pre}. It posits that many paragraphs in the training set documents contain intrinsic tasks such as sentiment analysis, and retrieves paragraphs of the same intrinsic task to compose in-context learning examples, but its retriever is trained on 37 downstream tasks, which are opposite to the ``intrinsic task'' nature, and limit the scale and diversity of the composed task data. In comparison, our BRD does not refer to any downstream tasks, and focuses on the contents of the training set texts, thus keeping more freedom in curating the task data. In addition, PICL aims to train the in-context learning ability, while BRD is for the model distillation. The intrinsic task data in PICL may not have task labels since the original paragraphs do not necessarily have both task queries and answers simultaneously, e.g., a sentiment expression paragraph may not explicitly states its positive or negative label for the sentiment analysis task. In comparison, BRD always gets education queries and responses.

\citet{zhang-etal-2023-babys} propose a similar intrinsic task pre-training approach that transforms fragmented sentences from babyLM training set into a cohesive paragraph \cite{warstadt2023papersbabylmchallenge}. Their task is quite challenging to accomplish since the sentences in the training set are sampled from diverse resources, and lack strong semantic ties with each other, resulting in the hardness of composing a cohesive paragraph. Such fiction data generation are different to our BRD approach, which generates solid basic education data on reading activities.

\section{Approach}

We use a subset of CommonCrawl (CC-100) corpus, which is usually included in LLMs pre-training, as the education resource to conduct the basic reading education. The whole education process contains two stages. In the first stage, for each sentence in the corpus, the teacher model is prompted to perform basic reading. In the second stage, we collect all basic reading behavior data to train the student model, and finally test the student model ability on various tasks.

\begin{table*}
    \centering
    \small
    \begin{tabular}{p{13cm}}
    \toprule
    Perform named entity recognition on a given sentence without recognizing personal pronouns in the input sentence as human names.   \\ 
    \textbf{Enter a sentence:}\newline Barack Obama was the 44th President of the United States.\newline \textbf{Output:} \newline In this sentence, "Barack Obama" is a person name entity, and "United States" is a geopolitical entity.
     \newline \textbf{Enter a sentence:}\newline I just bought a new MacBook Pro from Apple.\newline \textbf{Output:}\newline In this sentence, "Apple" is an organization name entity, and "MacBook Pro" is a product name entity.
     \newline \textbf{Enter a sentence:}\newline The Eiffel Tower is a famous landmark in Paris, France.\newline \textbf{Output:}\newline In this sentence, "Eiffel Tower" is a landmark name entity, and "Paris" and "France" are geopolitical entities. \\
    
    \textbf{Enter a sentence:}\newline Belmont Estate is on the market for \$63 million and boasts roughly 22,000 square feet of luxurious finishes and elaborate architecture on 1.28 acres.\newline \textbf{Output:} \\
    \hline
    In this sentence, "Belmont Estate" is a geographic entity, "63 million" is a numerical entity representing the price of the estate, and "1.28 acres" is a geographic entity representing the size of the estate.\\
    \bottomrule
      
    \end{tabular}
    \caption{The prompt for the teacher model to extract named entity information from an input sentence. Each example consists of a sentence and its named entity information. The response from the teacher model is listed in the bottom. }
    \label{tab:ner_prompt}
\end{table*}

\subsection{Basic Reading Behaviors of the Teacher Model}

We utilize the in-context learning ability of the teacher model to elicit its basic reading behaviors including named entity recognition, question raising and answering. Given the corpus, we set up a prompt template consisting of task description, task examples, and input sentence from the corpus. 

Table \ref{tab:ner_prompt} lists the named entity recognition prompt and the response from the teacher model. We can see that, given the few-shot examples including entities and their types, the teacher model responses with more detailed contents of the entities, such as the price or size of the entities, which are beneficial for educating the student model to grasp the important information contained in the input sentence. Table \ref{tab:qa_prompt} lists the question raising and answering prompt and the response from the teacher model. In the task instruction, question is constrained to be about the content, structure, or attitude of the input sentence. The question raising and answering embody the teacher model's reading ability, which is targeted to be transferred to the student model.

\subsection{Training the Student Model} \label{sec:brd_train}

The student model is initialized by a released smaller pre-trained language model. We continue training the student model based on the basic reading behavior data generated by the teacher model. To stabilize the training process, we mix the basic reading behavior data with the original sentences of the corpus to avoid the catastrophic forgetting of the pre-trained model.

Suppose we have a passage consisting of three sentences $s_1$, $s_2$, and $s_3$, we constitute the named entity recognition passage:
 $s_1$ <sep> NER($s_1$) <sep> $s_2$ <sep> NER($s_2$) <sep> $s_3$ <sep> NER($s_3$), 
where NER denotes the named entity recognition result of the teacher model for each sentence, and <sep> is the delimiter. Similarly, we constitute the question raising and answering passage: 
$s_1$ <sep> QRA($s_1$) <sep> $s_2$ <sep> QRA($s_2$) <sep> $s_3$ <sep> QRA($s_3$), where QRA denotes the question raising and answering result of the teacher model for each sentence. The original passage is formatted as $s_1$ <sep> $s_2$ <sep> $s_3$. We use passage instead of sentence to be consistent with the usual language model pre-training that utilizes long contexts.

In this way, we build all original passages, denoted as $D_{ORI}$, all named entity recognition passages, denoted as $D_{NER}$, and all question raising and answering passages, denoted as $D_{QRA}$. We mix them together to build the training set $D_{TRAIN}$, on which we train the student model to minimize the loss in an autoregressive manner:

\[ L = -\frac{1}{N}  \sum\limits_{i=1}^N \sum\limits_{t=1}^T {\rm log}P( y_{t} |y_{<t})\]

\noindent where $y$ is the passage with length $T$, and $N$ is the number of passages in $D_{TRAIN}$.

\begin{table*}
    \centering
    \small
    \begin{tabular}{p{13cm}}
    \toprule
    Ask a question to the input sentence, you can ask questions about the content, structure or attitude of the sentence, and then find the answer to the corresponding question in the original sentence. Output in the format "Question: Answer:".  \\
    \textbf{The sentence:} \newline In order to graduate with honors, he needed to maintain a high GPA throughout college.\newline
    \textbf{Question:} \newline What did he need to do in order to graduate with honors?\newline
    \textbf{Answer:} \newline Maintain a high GPA throughout college.
    \\
    \textbf{The sentence:} \newline Belmont Estate is on the market for \$63 million and boasts roughly 22,000 square feet of luxurious finishes and elaborate architecture on 1.28 acres.
    \\
    \hline
    \textbf{Question:} \newline How much does Belmont Estate cost?\newline
    \textbf{Answer:} \newline Belmont Estate costs \$63 million.
    \\
    \bottomrule
      
    \end{tabular}
    \caption{The prompt for the teacher model to perform question raising and answering on an input sentence. Question is limited to be about the input sentence. The response from the teacher model is listed in the bottom.}
    \label{tab:qa_prompt}
\end{table*}

\subsection{Testing} \label{sec:test}

For predicting the answers of the downstream tasks when testing the student model, we use the average of per-token log-probabilities of candidate answers as the scoring function for all downstream tasks:
\[ \bar{P} = \frac{1}{n} \sum_{i=1}^{n} \log P_i(y_i | x_{\rm prompt}) \]

\noindent where $x_{\rm prompt}$ denotes the input to the student model, $y$ denotes the candidate answer for $x_{\rm prompt}$, and $n$ is the total number of words in $y$. We select $y$ with the maximal $\bar{P}$ as the final answer for $x_{\rm prompt}$. This average computation is to cover tasks such as Google BIG-bench \footnote{\url{https://github.com/google/BIG-bench}} \cite{srivastava2023beyond}, whose candidate answers are phrases/sentences rather than single words.

\section{Experiment} \label{sec:exp}

We use the well-known LLM Vicuna-13B \footnote{\url{https://github.com/lm-sys/FastChat}} \cite{vicuna2023} as our teacher model due to its high efficiency in generating large volume of texts for teaching. We use XGLM-564M \cite{lin-etal-2022-shot} \footnote{\url{https://github.com/facebookresearch/fairseq/tree/main/examples/xglm}}, which is the smaller language model of the same decoder-only family, to initialize our student model. To compare the student model with larger model pre-trained on the same data origin, we also include XGLM-7.5B for comparison. In BRD, we use five million sentences from CC-100 corpus to collect the basic reading data generated by Vicuna-13B.  

\subsection{Baselines}

We consider three baselines in our experiments: 

\begin{itemize}

\item Knowledge distillation (KD): We use two KD models released in \citet{Gu2023KnowledgeDO}\footnote{\url{https://github.com/microsoft/LMOps/tree/main/minillm}} for the comparison. One is the standard KD (SKD) that uses teacher distribution to supervise the student model. The other is SOTA KD model MiniLLM that uses reverse Kullback-Leibler divergence for the distillation.

\item Task distillation \cite{wang-etal-2021-want-reduce,iliopoulos2022weighted}: The teacher model generates the answers given the downstream task inputs, and these generated pseudo answers are used to supervise the student model. 

\item Supervised Fine-tuning(SFT): Directly fine-tunes the student model on the downstream tasks supervised by the gold labels.

\end{itemize}

\subsection{Evaluation}

We adopt a spectrum of downstream tasks for the evaluation, including natural language inference (XNLI\cite{conneau2018xnli}, CB\cite{Marneffe2019TheCI}, RTE\cite{wang-etal-2018-glue}) , paraphrasing (PAWS-X\cite{zhang-etal-2019-paws}) , Boolean QA (BOOLQ\cite{clark-etal-2019-boolq}) , sentiment analysis (SST-2\cite{socher-etal-2013-recursive}), and Google BIG-bench\cite{srivastava2023beyond}. In Google BIG-bench tasks, we only consider multiple choice QA tasks which have the fixed answers easy for the evaluation, resulting in a total of 73 tasks. The results are evaluated by the accuracy of the predicted answers. The prompts for the downstream tasks are presented in the appendix \ref{appen:prompts}.

\subsection{Results}

The main results are grouped into three parts as shown in Table \ref{tab:main-results}. The top part presents the accuracies of the original models, including the teacher model Vicuna-13B, the student model XGLM-564M, the large model XGLM-7.5B which has same origin to the student model, plus an extension model XGLM-564M-FURTHER, which further trains the student model on the original one million passages from CC-100 corpus. The number of the further training steps is set 18,000.

The middle part and the bottom part list the accuracies of various distillation or fine-tuning approaches under two scenarios: without downstream task supervision and with downstream task supervision, respectively. The difference between the two scenarios is the availability of the downstream task gold answers.

\begin{table*}
    \centering
    \small
    \begin{adjustbox}{width=0.8\linewidth}
    \begin{tabular}{lcccccccc}
    \toprule
    & \multicolumn{8}{c}{Task} \\
    \cmidrule{2-8}
    \multicolumn{1}{c}{\textbf{Model}} & \textbf{XNLI} & \textbf{RTE}  & \textbf{CB} & \textbf{PAWS-X} & \textbf{BOOLQ} & \textbf{SST-2} & \textbf{BIG-bench-Avg} & \textbf{Average}\\
    \hline \hline
    Vicuna-13B & 59.1 & 78.3 & 71.4 & 62.9 & 84.3 & 81.5 &  35.6  & 67.6 \\
    XGLM-7.5B & 36.6 & 50.9 & 60.7 & 56.8 & 57.2 & 69.5 & 34.3  & 52.34 \\
    XGLM-564M & 35.5 & 46.2 & 53.6 & 51.3 & 51.2 & 63.9 & 34.0 & 48.0 \\
    XGLM-564M-FURTHER & 34.9 & 46.6 & 51.8 & 51.6 & 51.5 & 59.4 & 34.0  & 47.1 \\
    \hline \hline
     \multicolumn{9}{c}{Without Downstream Task Supervision} \\
    \hline
    SKD & 33.7 & 53.8 & 51.8 & 43.0 & 49.1 & 60.7  & 34.2 & 46.6 \\
    MiniLLM & 34.2 & \textbf{58.1} & \textbf{73.2} & 44.1 & 55.9 & 62.4 & 34.6 & 51.8 \\
    XGLM-BRD & \textbf{36.2} & 53.8 & 58.9 & \textbf{56.7} & \textbf{61.0} & \textbf{78.1} & \textbf{34.8} & \textbf{54.2} \\
    \hline
    TaskDistillation & 57.1 & 58.1 & 60.7 & 64.8 & 74.8 & 77.2  & 41.6$\dagger$  & 62.0  \\ 
    XGLM-BRD$^2$ &  \textbf{59.2} & \textbf{62.5} & \textbf{82.1} & 64.8 & \textbf{75.0} & \textbf{81.9}  & \textbf{44.1}$\dagger$  & \textbf{67.1}  \\
    \hline \hline
     \multicolumn{9}{c}{With Downstream Task Supervision} \\
    \hline
    SFT & 81.4 & 67.1 & 83.9 & \textbf{92.4} & 77.5 & 91.5 &  68.3$\dagger$ &  80.3   \\
    XGLM-BRD$^2$-SFT & \textbf{81.6} & \textbf{69.3} & \textbf{91.1} & 91.5 & \textbf{77.8} & \textbf{92.2} & \textbf{69.1}$\dagger$  & \textbf{81.8}  \\
    \bottomrule
    \end{tabular}
    \end{adjustbox}
    \caption{Main results of the teacher models, student models, and various distillation and fine-tuning approaches. Unless otherwise specified, the student models are all initialized by XGLM-564M. BIG-bench-Avg is the accuracy averaged over the 73 bench tasks($\dagger$ denotes the averaged accuracy on the reduced set of BIG-bench tasks), and detailed accuracies are reported in the appendix \ref{sec:bigbench}.}
    \label{tab:main-results}
\end{table*}

\paragraph{Results Without Downstream Task Supervision.} In this scenario, the downstream task gold answers are not available. It is further divided into two conditions. One is the blind test setting, in which any task training set data is NOT accessible. It is for applications of the student model on fairly new tasks. The other is the relaxed test setting, in which only the training set input data (without gold answers) are accessible. It is for applications on tasks that manual labeling for the training set input data is not available. 

\begin{itemize}

\item
In the blind test setting, we compare our XGLM-BRD with the two released KD works: SKD and MiniLLM. In the multiple student models of the two KD works, we select their GPT-2 760M version student models for the comparison due to the similar model size. The results show that XGLM-BRD performs significantly better than SKD and MiniLLM in most tasks, demonstrating that XGLM-BRD has better generalization ability to various unseen tasks. We also combine BRD with the two KD works, and the results are listed in the orthogonal analysis section \ref{sec:orthogonal} and Table \ref{tab:orthogonal}.

Regarding the comparison between XGLM-BRD and XGLM-564M, BRD significantly improves the performance of the small model, indicating that basic reading education does enhance the ability of the small model. Moreover, XGLM-564M-FURTHER performs much worse than XGLM-BRD, revealing that only using the original passages for further training does not yield enhancements and may even leads to decreases in some tasks. It is the basic reading data for further training that advance the student model. XGLM-BRD also approaches or even surpasses XGLM-7.5B, which is 15x bigger, on the downstream tasks. There is still a gap between XGLM-BRD and the teacher model Vicuna-13B, but this gap is significantly reduced or disappeared when we conduct relaxed test.

In some tasks such as CB and RTE, there is comparison variance between MiniLLM and BRD. This may be because  MiniLLM was instruction tuned on databricks-dolly-15k, which is a task instruction following dataset that may cover some traditional downstream tasks such as CB and RTE, leading to the performance better than our BRD approach. In comparison, our BRD model was not instruction tuned, but was trained on general domain to improve its general reading ability, leading to the overall performance superior to MiniLLM on wide variety of 79 downstream tasks.

\begin{table*}[]
\centering
\small
\begin{tabular}{l | l | lllllll}
\toprule
\multirow{2}{*}{Model}      & \multirow{2}{*}{Approach} & \multicolumn{7}{c}{Task}                           \\
                            &                         & XNLI & RTE  & CB   & PAWS-X & BOOLQ & SST-2 & Average  \\
\hline \hline
\multirow{4}{*}{GPT-2 120M} & SKD                     & 35.9 & 44.4 & 57.1 & 52.1   & 47.6  & 68.8  & 51.0 \\
                            & \ \ \ \ +BRD                    & \textbf{37.0} & \textbf{54.5} & \textbf{66.1} & \textbf{56.8}   & \textbf{62.2}  & \textbf{76.9}  & \textbf{58.9} \\ 
                            & MiniLLM                 & \textbf{35.9} & 48.0   & 67.9 & \textbf{57.0}   & 53.1  & 53.8  & 52.6 \\
                            & \ \ \ \ +BRD                    & 35.1 & \textbf{51.6} & \textbf{71.4} & 48.5   & \textbf{60.9}  & \textbf{79.1}  & \textbf{57.8} \\
\hline
\multirow{4}{*}{GPT-2 340M} & SKD                     & 33.6 & 46.9 & 51.8 & \textbf{54.2}   & 48.8  & 63.3  & 49.8 \\
                            & \ \ \ \ +BRD                    & \textbf{34.2} & \textbf{55.2} & \textbf{67.9} & 53.8   & \textbf{64.6}  & \textbf{78.1}  & \textbf{59.0} \\
                            & MiniLLM                 & \textbf{32.8} & 46.6 & 50.0 & \textbf{57.0}   & 56.9  & 55.3  & 49.0 \\
                            & \ \ \ \ +BRD                    & 32.7 & \textbf{56.3} & \textbf{67.9} & 53.4   & \textbf{64.1}  & \textbf{74.4}  & \textbf{58.1} \\
\hline
\multirow{4}{*}{GPT-2 760M} & SKD                     & 33.7 & \textbf{53.8} & 51.8 & 43.0   & 49.1  & 60.7  & 48.7 \\
                            & \ \ \ \ +BRD                    & \textbf{34.7} & 52.3 & \textbf{64.3} & \textbf{56.4}   & \textbf{62.1}  & \textbf{68.5}  & \textbf{56.7} \\
                            & MiniLLM                 & 34.2 & \textbf{58.1} & 73.2 & 44.1   & 55.9  & 62.4  & 54.7 \\
                            & \ \ \ \ +BRD                    & \textbf{35.2} & 52.0 & \textbf{76.8} & \textbf{50.8}   & \textbf{60.5}  & \textbf{67.3}  & \textbf{57.1}  \\
\hline
\multirow{2}{*}{XGLM-564M} & TaskDistillation & 57.1 & 58.1 & 60.7 & \textbf{64.8} & \textbf{74.8} & 77.2  & 65.5  \\
                            & \ \ \ \ +BRD                    &  \textbf{58.1} & \textbf{61.0} & \textbf{71.4} & 63.1 & 74.4 & \textbf{81.1} & \textbf{68.2}  \\
\bottomrule
\end{tabular}
\caption{Results of combining BRD with various distillation approaches. The models for initializing the student models are listed in the model column.}
\label{tab:orthogonal}
\end{table*}

\begin{table*}[]
\centering
\small
\begin{tabular}{l | lllllll}
\toprule
                            & XNLI                 & RTE                  & CB                   & PAWS-X             & BOOLQ              & SST-2                &    BIG-bench-Avg      \\
          \hline
XGLM-564M       & 461.7                & \textbf{64.8}       & 13.2                 & 1065.6                & 1098.7               & 55.6                  &     145.1                       \\
XGLM-BRD        & \textbf{407.5}    & 66.2                   & \textbf{12.8}     & \textbf{892.4}      & \textbf{1001.1}    & \textbf{37.0}     &     \textbf{112.2}          \\
\bottomrule
\end{tabular}
\caption{Cross entropy between the distributions of the teacher model and small models. The lower the better for measuring the consistency.}
\label{tab:entropy}
\end{table*}

\item
In the relaxed test setting, we compare our BRD with the task distillation approach(TaskDistillation in Table \ref{tab:main-results}), which uses the teacher model to generate pseudo answers on the task training set for supervising the student model XGLM-564M. Because BIG-bench tasks do not divide training, tuning, and test sets, we only consider tasks each of whom has more than 2K instances in the relaxed test, and finally select tasks that rank top-5 according to the number of instances as the reduced set of BIG-bench tasks(denoted by $\dagger$ in Table \ref{tab:main-results}). For each task, we save ten percent of instances as test set, ten percent of instances as tuning set, and other instances as training set. Our approach in this setting uses BRD twice, that is, on the general data we conduct BRD to obtain the student model XGLM-BRD, then on the downstream task data, we conduct BRD again to continual training the new student model, denoted as XGLM-BRD$^2$. The results show that the task distillation approach establishes a strong baseline that significantly outperforms both XGLM-564M and KD models, demonstrating that even pseudo answers can supervise the student model to perform well on the downstream tasks. When BRD is introduced into this process, the improvement is even more pronounced by XGLM-BRD$^2$. The comparison between basing on XGLM-BRD and basing on XGLM-564M in this test setting is presented in Appendix \ref{appen:fur}.

When comparing XGLM-BRD$^2$ with the teacher model Vicuna-13B, it shows that XGLM-BRD$^2$ outperforms Vicuna-13B in some tasks, and in the other tasks, the performance gap is significantly reduced. This comparison proves the effectiveness of BRD, that leads to comparable or superior performance to the 26x bigger teacher model. 

\end{itemize}

\paragraph{Results With Downstream Task Supervision.} In this scenario, the downstream task gold answers are available. We compare BRD with SFT, which fine-tunes the student model XGLM-564M based on the task supervision data. Table \ref{tab:main-results} shows that with the gold supervision, SFT significantly improves the ability of the student model, and beats the 26x bigger model Vicuna-13B with a large margin in certain tasks. In comparison to this strong baseline, we conduct BRD on the task data with gold answer, then continue training XGLM-BRD on the data mixing this basic reading data and the original task data. The trained model is denoted as XGLM-BRD$^2$-SFT. The results show that XGLM-BRD$^2$-SFT surpasses SFT in most tasks, demonstrating the effectiveness of the basic reading education for the student model when the downstream task supervision is available.

\section{Analysis}

\subsection{Orthogonality of BRD to Knowledge Distillation and Task Distillation} \label{sec:orthogonal}

Since BRD focuses on basic reading education for the student model without referring to any implicit model features or downstream tasks, it is orthogonal to either knowledge distillation or task distillation. So, we combine BRD with knowledge distillation by further training the student model of knowledge distillation on our general basic reading data, or combine BRD with task distillation  by further training the student model of task distillation on the basic reading data of the downstream tasks. Table \ref{tab:orthogonal} lists the combination results.

It shows that combining BRD in most cases significantly improves the performances of the two distillation approaches, which proves the orthogonality of BRD to either knowledge distillation or task distillation.

\begin{table*}
    \centering
    \small
    \begin{adjustbox}{width=0.8\linewidth}
    \begin{tabular}{lccccccc}
    \toprule
    & \multicolumn{7}{c}{Task} \\
    \cmidrule{2-7}
    \multicolumn{1}{c}{\textbf{Model}} & \textbf{XNLI} & \textbf{RTE}  & \textbf{CB} & \textbf{PAWS-X} & \textbf{BOOLQ} & \textbf{SST-2} & \textbf{Average}\\
    \hline \hline
    Llama3.1-8B & 49.6 & 79.8 & 73.2 & 66.0 &  84.4  & 92.0 & 74.2  \\
    Vicuna-13B & 59.1 & 78.3 & 71.4 & 62.9 & 84.3 & 81.5  & 72.9  \\
    \hline
    XGLM-564M & \textbf{35.5} & 46.2 & 53.6 & 51.3 & 51.2 & 63.9 & 50.3 \\
    XGLM-BRD-Llama  & 32.0  & \textbf{57.0}  &  \textbf{69.6}  &  52.8 & 58.5  & \textbf{71.9}  & \textbf{57.0}   \\
    XGLM-BRD-Vicuna & 33.7 & 56.0  & 60.7  &  \textbf{54.4}  &  \textbf{59.8}  & 69.0 & 55.6  \\
    \bottomrule
    \end{tabular}
    \end{adjustbox}
    \caption{Comparison between using Llama3.1-8B and Vicuna-13B as the teacher models on the reduced dataset of one million sentences. }
    \label{tab:llama-results}
\end{table*}

\subsection{Effectiveness Verification Based on Cross Entropy Evaluation}

BRD educates the student model via explicit basic reading behaviors. We study if such education can effectively influence the probability distribution of the student model. We compute the cross entropy, which is often used to measure the consistency between the teacher distribution and the student distribution, for the teacher model Vicuna-13B and the student model XGLM-BRD: 

\[ - \sum\limits_{i=1}^N p(y) {\rm log}q(y')\]

\noindent where $p$ is the teacher model probability, $q$ is the student model probability, $y$ and $y'$ are subword sequences of the same text according to the teacher model and the student model, respectively. $N$ is the number of the texts. Since $y$ and $y'$ have different lengths, we set $p$ and $q$ as sequence-level probabilities averaged over $y$ and $y'$, respectively.
We use the instances from the downstream tasks for computing the cross entropy. For the considered 73 tasks in BIG-bench, we randomly choose 1K instances from each task for the computation, and report the cross entropy averaged over the tasks. We include the original XGLM-564M to compute $q$ for comparison.

Table \ref{tab:entropy} shows the comparison result in the blind test. Lower cross entropy means better consistency between the teacher model and the student model. It shows that on most downstream tasks, XGLM-BRD approaches more closer to the teacher model than the original XGLM-564M does, demonstrating significant advantage in shaping the student model probability distribution towards that of the teacher model.

\subsection{Extension to Using Llama as The Teacher Model}

In addition to using Vicuna-13B as the teacher model, we explore setting Llama3.1-8B-Instruct(abbreviated as Llama3.1-8B) as the teacher model in the blind test, and compare both based on a subset of the data used in section \ref{sec:exp}. The subset consists of one million sentences, on which we conduct BRD. For each teacher model, we use XGLM-564M as the student model.

Table \ref{tab:llama-results} lists the comparison results. XGLM-BRD-Llama/Vicuna denotes the model distilled from Llama3.1-8B/Vicuna-13B. It shows that BRD is effective for using either Llama3.1-8B or Vicuna-13B as the teacher model, achieving significant improvements over XGLM-564M in most tasks. Regarding the comparison between Llama3.1-8B and Vicuna-13B as the teacher model, Llama3.1-8B averagely performs better than Vicuna-13B as the teacher in BRD.

\subsection{Comparison to Sentence Level Training}

In building the BRD training data presented in section \ref{sec:brd_train}, we divide a passage into sentences, then annotate each sentence with basic reading behaviors by using the teacher model, and finally compose all sentences and their annotations into a passage according to the original sentence order. To check whether this passage level training has the positive effect, we abandon the last composing step and leave the sentences and their annotations unordered. Then we conduct the sentence level training on this dataset to compare with the passage level training. Table \ref{tab:passage_sentence} presents the comparison result. 

It shows that the sentence level training generally performs worse than the passage level training. Since the downstream tasks are mostly the tasks with multiple sentences as input, the passage level training is more suitable for the downstream tasks than the sentence level training due to its multiple sentence training nature.

\begin{table}[t!]
    \centering
    \begin{adjustbox}{width=1\linewidth}
    \begin{tabular}{r ccccccc}
    \toprule
    & \multicolumn{6}{c}{Tasks} \\
    \cmidrule{2-7}
    \multicolumn{1}{c}{\textbf{ }} & \textbf{XNLI} & \textbf{RTE}  & \textbf{CB} & \textbf{PAWS-X} & \textbf{BOOLQ} & \textbf{SST-2} & \textbf{Avg}\\
    \midrule
    XGLM-564M & 35.5 & 46.2 & 53.6 & 51.3 & 51.2 & 63.9 &  50.3  \\
    PassageLevel & \textbf{36.2} & 53.8 & \textbf{58.9} & \textbf{56.7} &  \textbf{61.0} &   \textbf{78.1}  &  \textbf{57.5}  \\
    SentenceLevel & 34.1 & \textbf{55.6}  &  55.4  &  53.6  &  58.9  &  76.0  &  55.6    \\
    \bottomrule
    \end{tabular}
    \end{adjustbox}
    \caption{The comparison between the passage level training and the sentence level training evaluated by the blind test.}
    \label{tab:passage_sentence}
\end{table}

\subsection{Ablation of Different Basic Reading Behaviors.} We test the contribution of the different basic reading behaviors by deleting either NER or QRA data of the downstream tasks in training XGLM-BRD$^2$. Table \ref{tab:different-datasets-results} lists the ablation results in the relaxed test. It shows that deleting the QRA data impacts the performance more significantly than deleting the NER data in most tasks. QRA focuses on the sentence understanding, thus contributing more in the basic reading education.

The coordination between NER and QRA is related to the multi-task learning \cite{10.1145/3663363} that boosts the model ability through training on multiple tasks with potential generalization to other tasks. Different to the multi-task learning that predefines a fixed set of tasks, BRD focuses only on the basic reading education that has flexible contents changing from sentence to sentence. This flexibility‌ empowers the distilled model to perform well on various downstream tasks.

\begin{table}[t!]
    \centering
    \begin{adjustbox}{width=1\linewidth}
    \begin{tabular}{r ccccccc}
    \toprule
    & \multicolumn{6}{c}{Tasks} \\
    \cmidrule{2-7}
    \multicolumn{1}{c}{\textbf{ }} & \textbf{XNLI} & \textbf{RTE}  & \textbf{CB} & \textbf{PAWS-X} & \textbf{BOOLQ} & \textbf{SST-2} & \textbf{Avg}\\
    \midrule
    XGLM-BRD$^2$ & 59.2 & 62.5 & 82.1 & 64.8 & 75.0 & 81.9  & 70.9  \\
     $-$NER & 58.0 & 61.4 & 71.4 & 64.1 & 74.3 &  81.4 & 68.4  \\
     $-$QRA & 58.3 & 61.0 & 67.9 & 63.9 & 74.9 & 80.5 & 67.8  \\
    \bottomrule
    \end{tabular}
    \end{adjustbox}
    \caption{The effects of deleting different basic reading behaviors for XGLM-BRD$^2$ in the relaxed test.}
    \label{tab:different-datasets-results}
\end{table}


\subsection{The Impact of BRD Data Size} 
\begin{figure}[t!]
    \centering

    \includegraphics[width=0.5\textwidth]{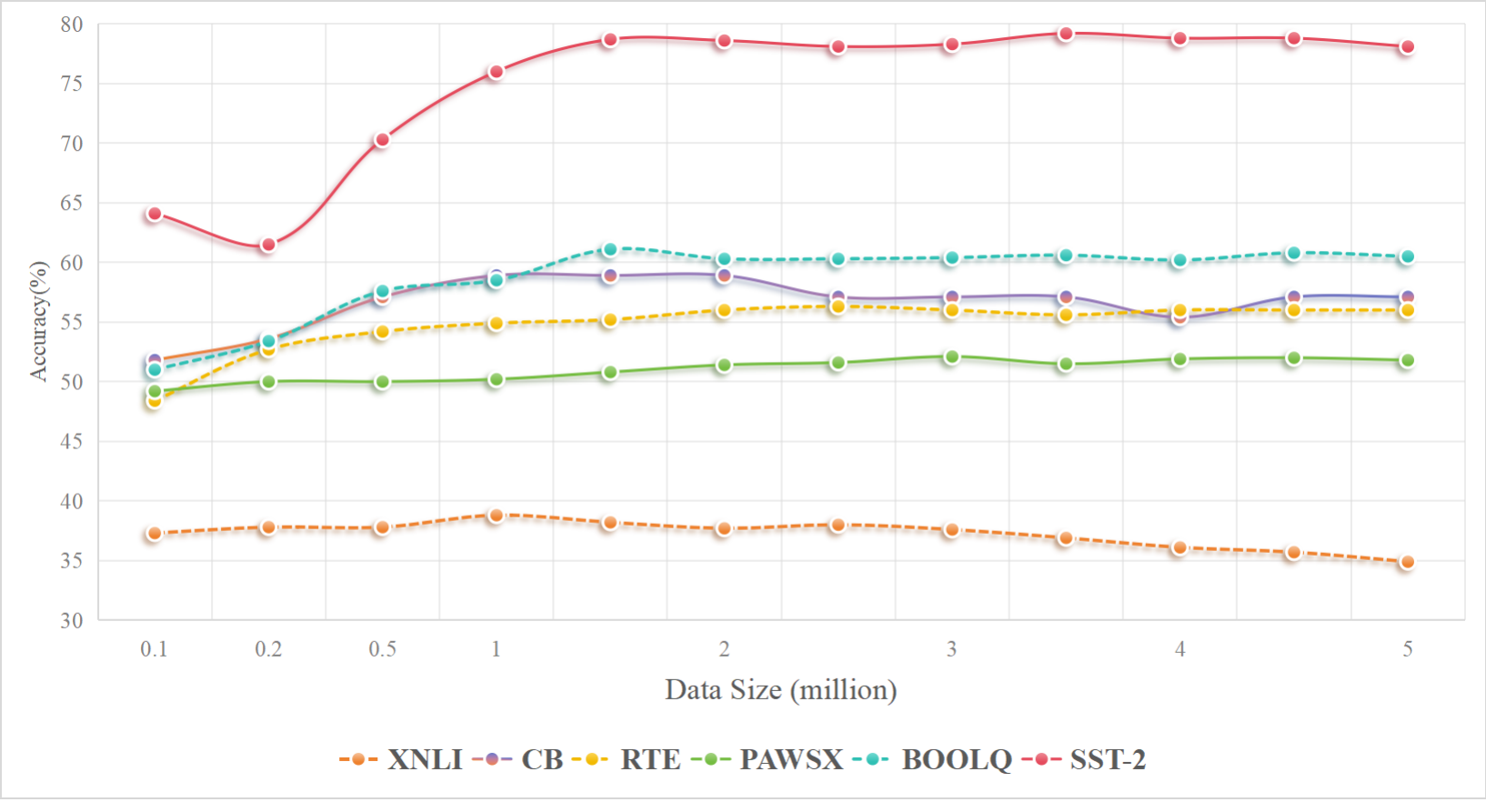}

    \caption{The performance curve along with different BRD data sizes (in million passages).}
    \label{fig:data_scale}
\end{figure}

We investigate how performance varies along with different BRD data sizes in the blind test. Figure \ref{fig:data_scale} shows the curve. Most tasks exhibit a steady improvement as BRD data gets bigger, and the performance plateaued when BRD data size arrives at more than one million passages.

\section{Conclusion}

In this paper, we propose to distill the basic reading abilities of LLMs into small models. In particular, we collect basic reading behaviors of LLMs such as NER or question raising and answering about parts of an input text at first, then we train small models based on the collected behaviors. Through such basic education on general texts, the small models are well educated to perform better on the downstream tasks. Experiments on various tasks including language inference benchmarks and Google Big-Bench tasks show that the small models after such distillation can surpass or perform comparable to LLMs that are 20x bigger. Verification by the cross entropy shows that such basic reading education can drive small model distribution closer to its teacher model distribution, leading to better performances than non-educated ones. Analysis also reveals that BRD has orthogonality to either knowledge distillation or task distillation.

\section{Acknowledgements}

We would like to thank the anonymous reviewers for the helpful comments. This work was supported by National Natural Science Foundation of China (Grant No. 62261160648, 62276179), and Project Funded by the Priority Academic Program Development of Jiangsu Higher Education Institutions.

\section*{Limitations}

In the distillation approach, we acknowledge certain limitations in the coverage of language models. We only use Vicuna-13B as our teacher model due to its high efficiency in generating large volume of texts. Calling the recent proprietary LLMs through API or using larger released LLMs incurs high cost in time and deployment for the massive distillation. It represents an area for potential future exploration to provide a more comprehensive understanding of using larger LLMs as the teacher model for the distillation.

\section*{Ethics Statement}

We honor the Code of Ethics. We do not use any private data or non-public information in this work. The language models used in this paper are freely downloadable from web. The corpus for generating basic reading behaviors by the teacher model is commonly used in most LLMs pretraining, and is freely released. The downstream task data are also freely downloadable from web. The distillation process does not involve any personally sensitive information.

\bibliography{acl_latex}

\appendix

\section{Appendix}
\label{sec:appendix}

\subsection{Training Configuration}

The student model is trained with learning rate = 0.0003, batch size = 8, and max input length = 2048, for a maximum of 40000 steps. We save the model every 1000 steps. Four A100 GPUs are used in both the data synthesis and the distillation training.

\subsection{Prompts for The Downstream Tasks}  \label{appen:prompts}

The prompt templates for the downstream tasks are listed in table \ref{tab:task_prompt}. For the 73 tasks in BIG-bench, we follow the general instruct with the task prefix and input as the prompt. 

\begin{table*}
    \centering
    \small
    \begin{tabular}{c|p{6cm}|c}
    \hline
    \multirow{2}{*}{\textbf{Task}} & \multirow{2}{*}{\textbf{Template}} & \multirow{2}{*}{\textbf{Candidate Answers}} \\
    & & \\
    \hline
    \multirow{4}{*}{XNLI} & \{premise\} & \multirow{4}{*}{Yes | No | Maybe}\\
    & Question: Does this imply that "\{hypothesis\}"? Yes, no or maybe? & \\
    & Answer: & \\
    \hline
    \multirow{2}{*}{RTE} & Question: Can we infer that "\{hypothesis\}" ? & \\
    \multirow{2}{*}{CB} & \{premise\} & \multirow{3}{*}{Yes | No | Maybe}\\
    & Answer: & \\
    \hline
    \multirow{5}{*}{PAWS-X} & Sentence 1: \{sentence1\} & \multirow{5}{*}{Yes | No} \\
    & Sentence 2: \{sentence2\} \\
    & Question: Do Sentence 1 and Sentence 2 express the same meaning? & \\
    & Answer: & \\
    \hline
    & \{passage\} & \\
    BOOLQ & Question: \{question\} & Yes |  No \\
    & Answer: & \\ 
    \hline
    \multirow{4}{*}{SST-2} & Question: Does the following sentence have a positive or negative sentiment? & \multirow{4}{*}{positive | negative} \\
    & Sentence: \{sentence\} & \\
    & Answer: & \\
    \hline

    \end{tabular}
    \caption{The prompt templates for the downstream tasks.}
    \label{tab:task_prompt}
\end{table*}

\subsection{Detailed Accuracies on BIG-bench} \label{sec:bigbench}

\begin{figure*}[t!]
    \centering

    \includegraphics[width=0.9\textwidth]{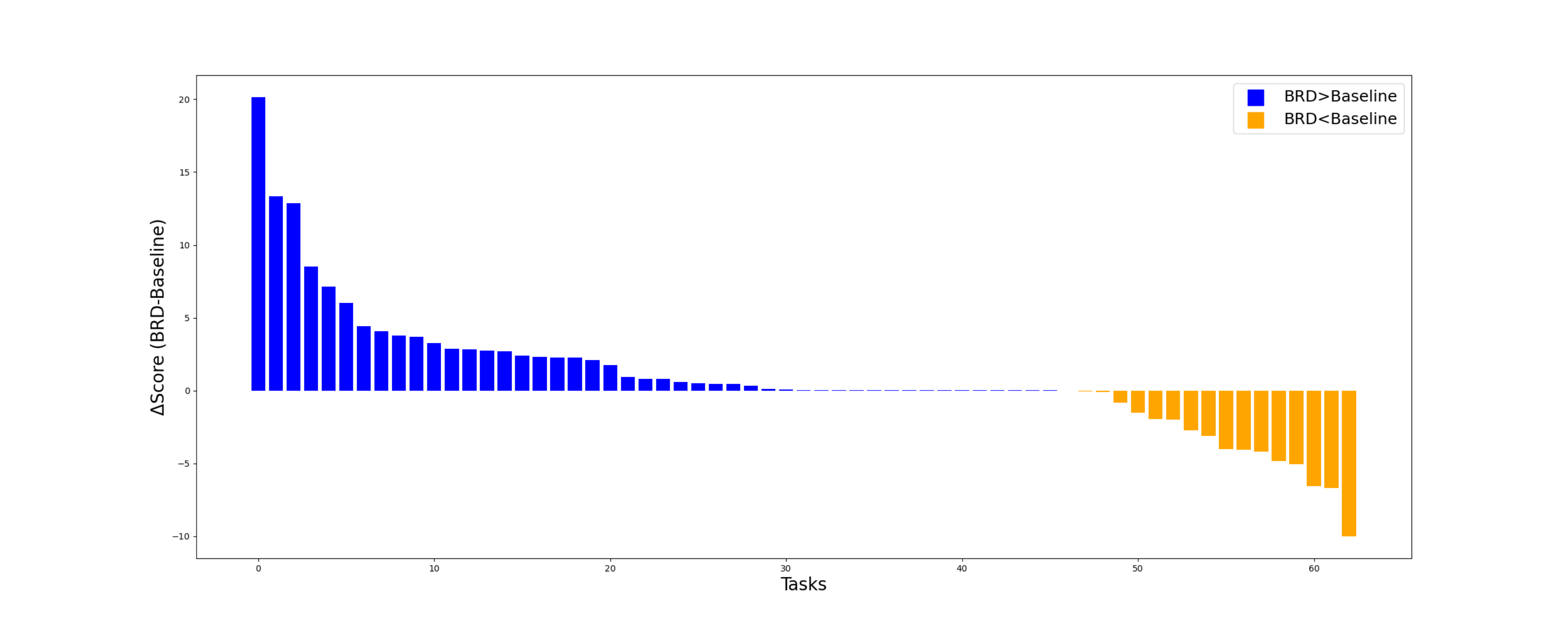}

    \caption{The accuracy comparison between BRD and the baseline on BIG-bench tasks.}
    \label{fig:bigbench}
\end{figure*}

Figure \ref{fig:bigbench} presents the accuracy comparison between the distilled model XGLM-BRD and the baseline model XGLM-564M on the 73 tasks in BIG-bench in the blind test. The results of ten tied tasks are not listed in the figure. It shows that XGLM-BRD improves the performances on about 2/3 tasks, demonstrating better ability generalized to a wide range of tasks than the baseline.

In the relaxed test and the setting with the downstream task supervision, we select tasks that rank top-5 according to the number of instances for the sufficiency consideration of dividing training, tuning, and test sets on these tasks. The five tasks are movie dialogue, formal fallacies and syllogisms with negation, Shakespeare dialogue, VitaminC, and WinoWhy. Table \ref{tab:bigbench-results} presents the results on this reduced BIG-bench set. It shows that BRD models perform superior to the corresponding baselines no matter the supervisions are available or not.

\begin{table*}
    \centering
    \small
    \begin{adjustbox}{width=0.8\linewidth}
    \begin{tabular}{lccccccc}
    \toprule
    & \multicolumn{6}{c}{Task} \\
    \cmidrule{2-6}
    \multicolumn{1}{c}{\textbf{Model}} & \textbf{MovieDialog} & \textbf{FormalFallacies}  & \textbf{ShakespeareDialogue} & \textbf{VitaminC} & \textbf{WinoWhy} &  \textbf{Average}\\
    \hline \hline
    \multicolumn{6}{c}{Relaxed Test} \\
    \hline
    TaskDistillation & 46.7 & 50.0 & 42.2 & 13.6 & 55.2 & 41.6   \\ 
    XGLM-BRD$^2$ &  \textbf{50.1} & 50.0 & \textbf{49.8} & 13.8 & \textbf{56.9} &   \textbf{44.1}  \\
    \hline \hline
     \multicolumn{6}{c}{With Downstream Task Supervision} \\
    \hline
    SFT & 69.2 & \textbf{69.9} & 69.3 & 55.9 & 76.9 &  68.3    \\
    XGLM-BRD$^2$-SFT & \textbf{70.6} & 69.3 & \textbf{70.2} & \textbf{56.0} & \textbf{79.1} &  \textbf{69.1}     \\
    \bottomrule
    \end{tabular}
    \end{adjustbox}
    \caption{Results on the reduced BIG-bench set.}
    \label{tab:bigbench-results}
\end{table*}

\subsection{Basing on XGLM-BRD is better than basing on XGLM-564M} \label{appen:fur}

In the relaxed test setting, further BRD in training XGLM-BRD$^2$ on the downstream tasks is based on XGLM-BRD. We also test further BRD based on XGLM-564M, which is denoted as XGLM-564M-FBRD. Table \ref{tab:MIX-results} lists the comparison result. It shows that XGLM-BRD$^2$ generally outperforms XGLM-564M-FBRD across various downstream tasks, highlighting that basing on XGLM-BRD is more effective. These results emphasize the importance of BRD as a prerequisite step in improving the adaptability and efficacy of models in downstream applications.

\begin{table}
    \centering
    \begin{adjustbox}{width=1\linewidth}
    \begin{tabular}{lcccccc}
    \toprule
    \cmidrule{2-7}
    \multicolumn{1}{c}{ } & \textbf{XNLI} & \textbf{RTE}  & \textbf{CB} & \textbf{PAWS-X} & \textbf{BOOLQ} & \textbf{SST-2}\\
    \midrule
    XGLM-564M-FBRD & 58.1 & 61.0 & 71.4 & 63.1 & 74.4 & 81.1 \\
    XGLM-BRD$^2$ & \textbf{59.2} & \textbf{62.5} & \textbf{82.1} & \textbf{64.8} & \textbf{75.0} & \textbf{81.9} \\
    \bottomrule
    \end{tabular}
    \end{adjustbox}
    \caption{The comparison between basing on XGLM-BRD and basing on XGLM-564M for further BRD on the downstream tasks.}
    \label{tab:MIX-results}
\end{table}

\subsection{Layer-wise Probing}

Inserting probes can reveal the interpretable aspects hidden in the neural networks \cite{belinkov-2022-probing}. We insert probes layer-wisely to check the efficacy of the distilled student model. In particular, for each downstream task, we extract the representation by averaging vectors per layer for each sentence in the training set, and train the probing classifier per layer based on the representation. The training loss is the regularized cross-entropy loss of the task prediction against the true label of the sentence. Through inserting probes layer-wisely, we can check how well each layer prepares for the downstream tasks. 

\begin{figure*}[!t]
    \centering

    \includegraphics[width=0.3\textwidth]{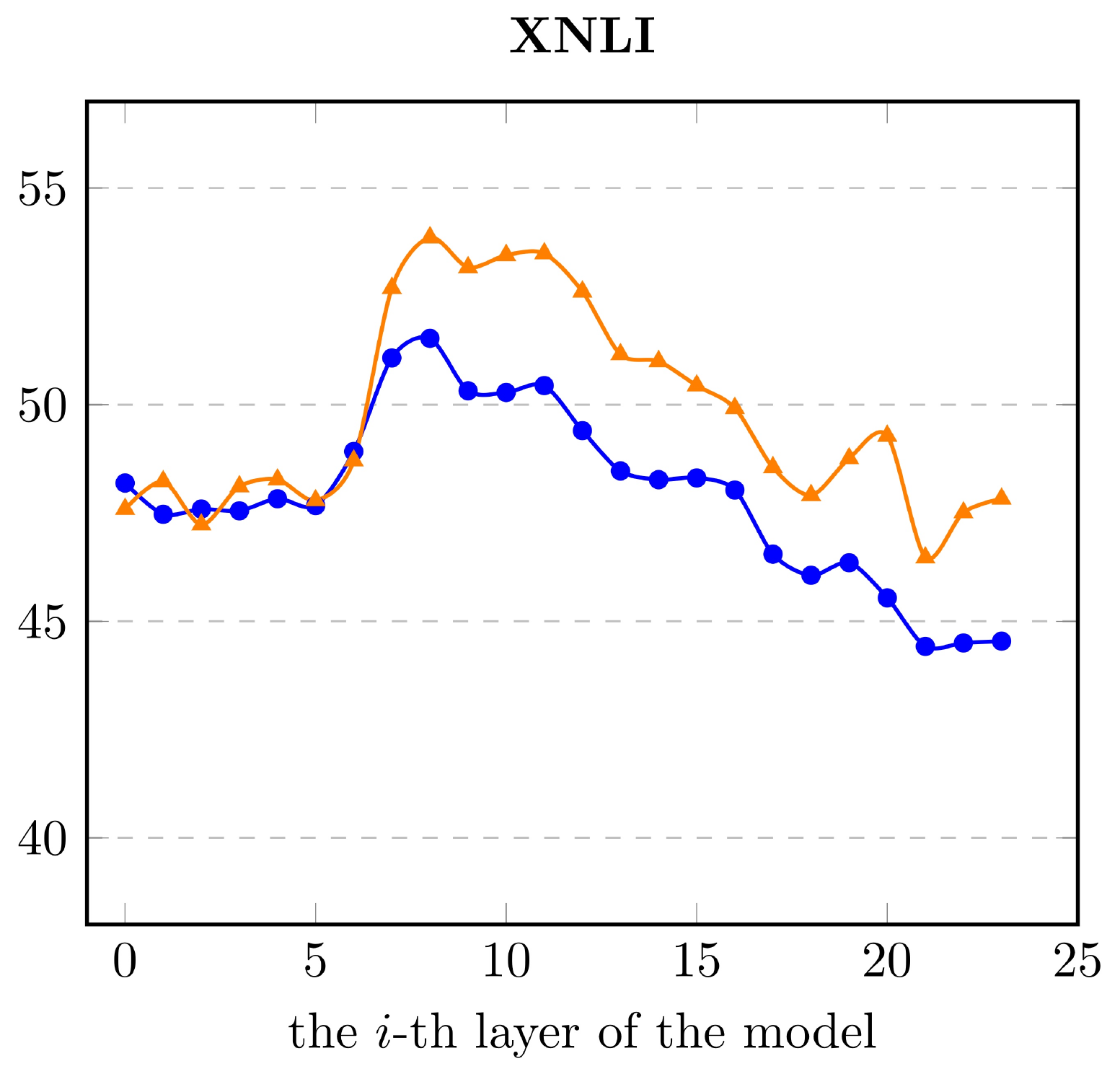}
    \includegraphics[width=0.3\textwidth]{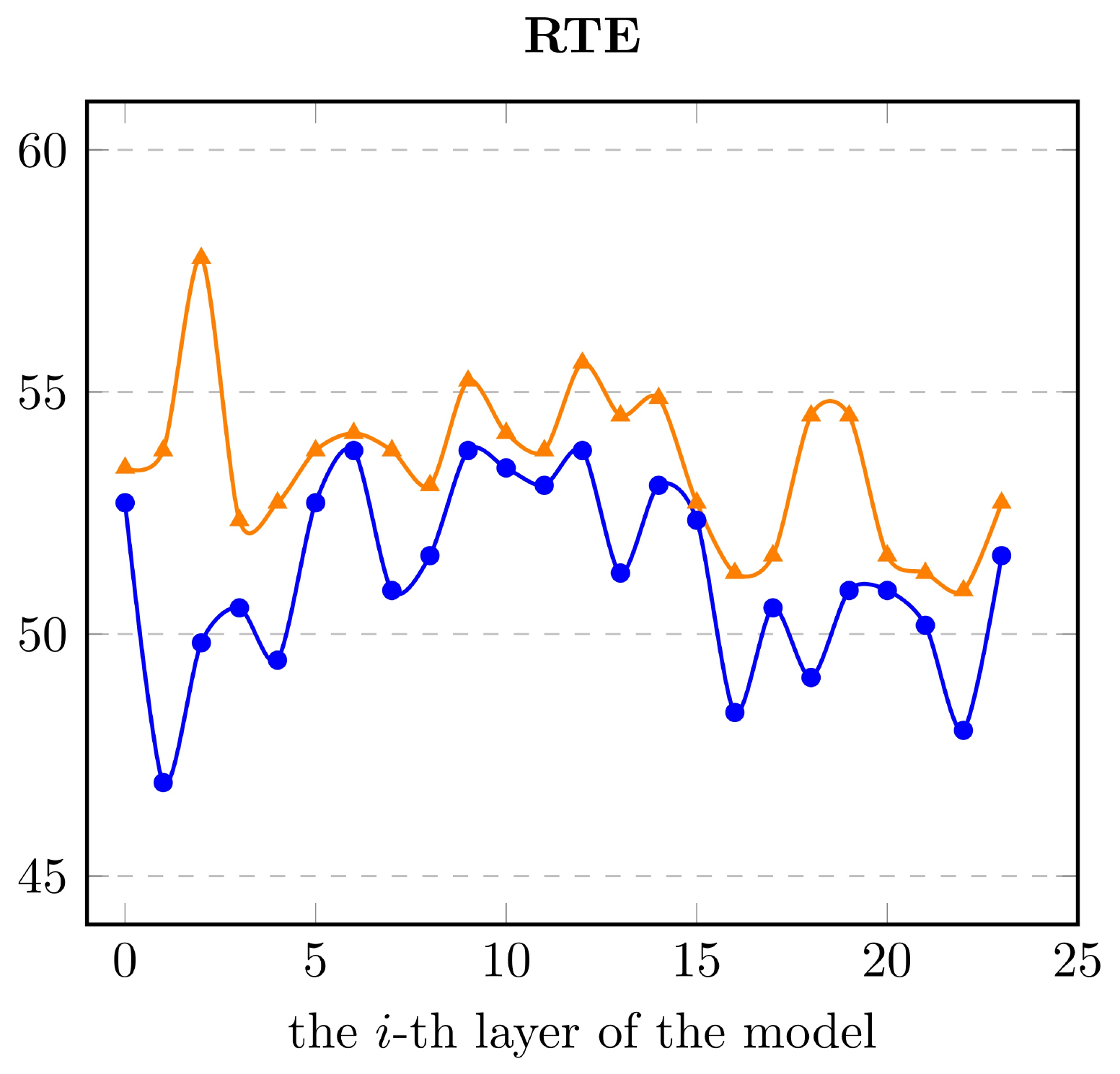}
    \includegraphics[width=0.3\textwidth]{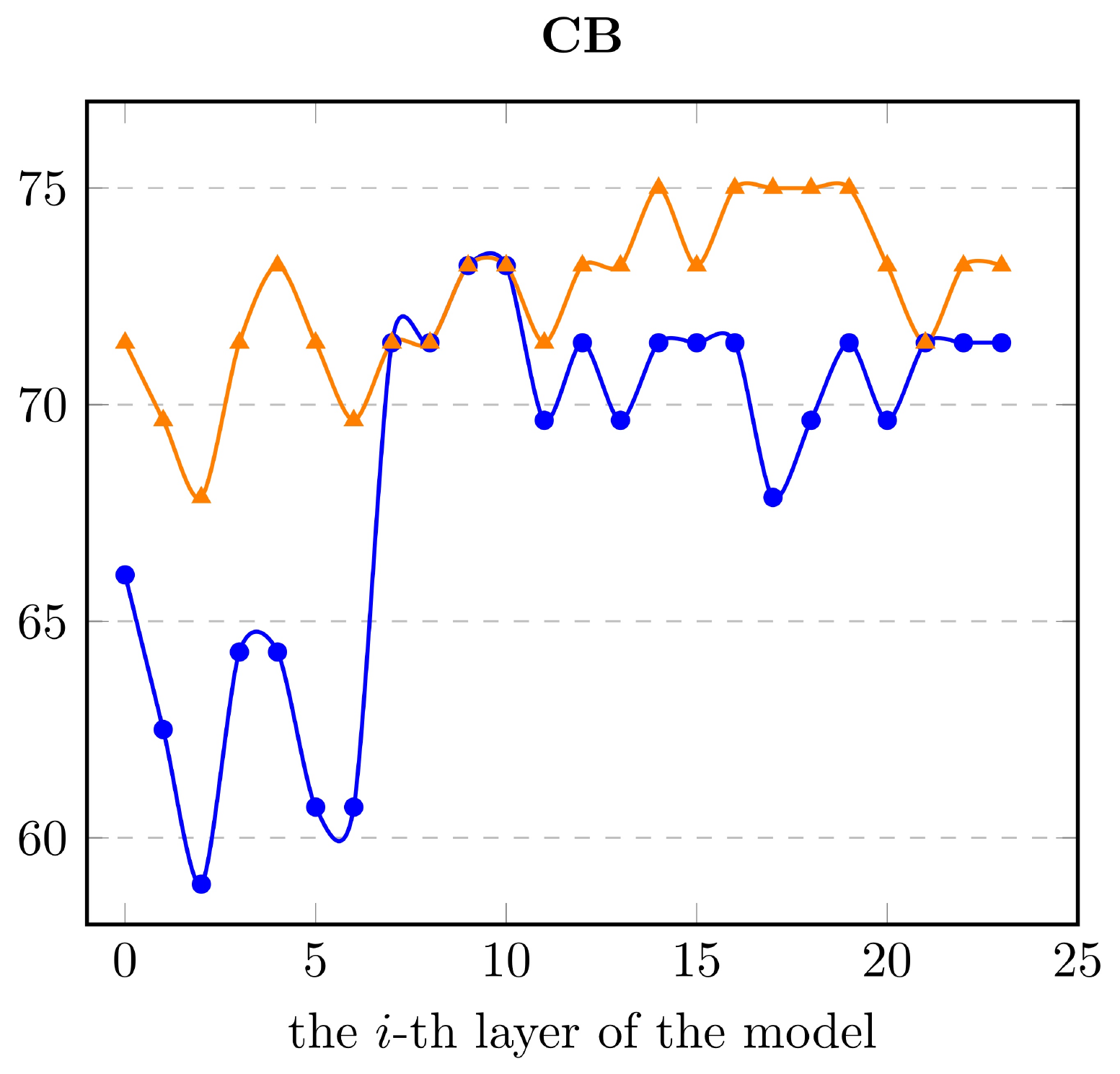}
    
    \includegraphics[width=0.3\textwidth]{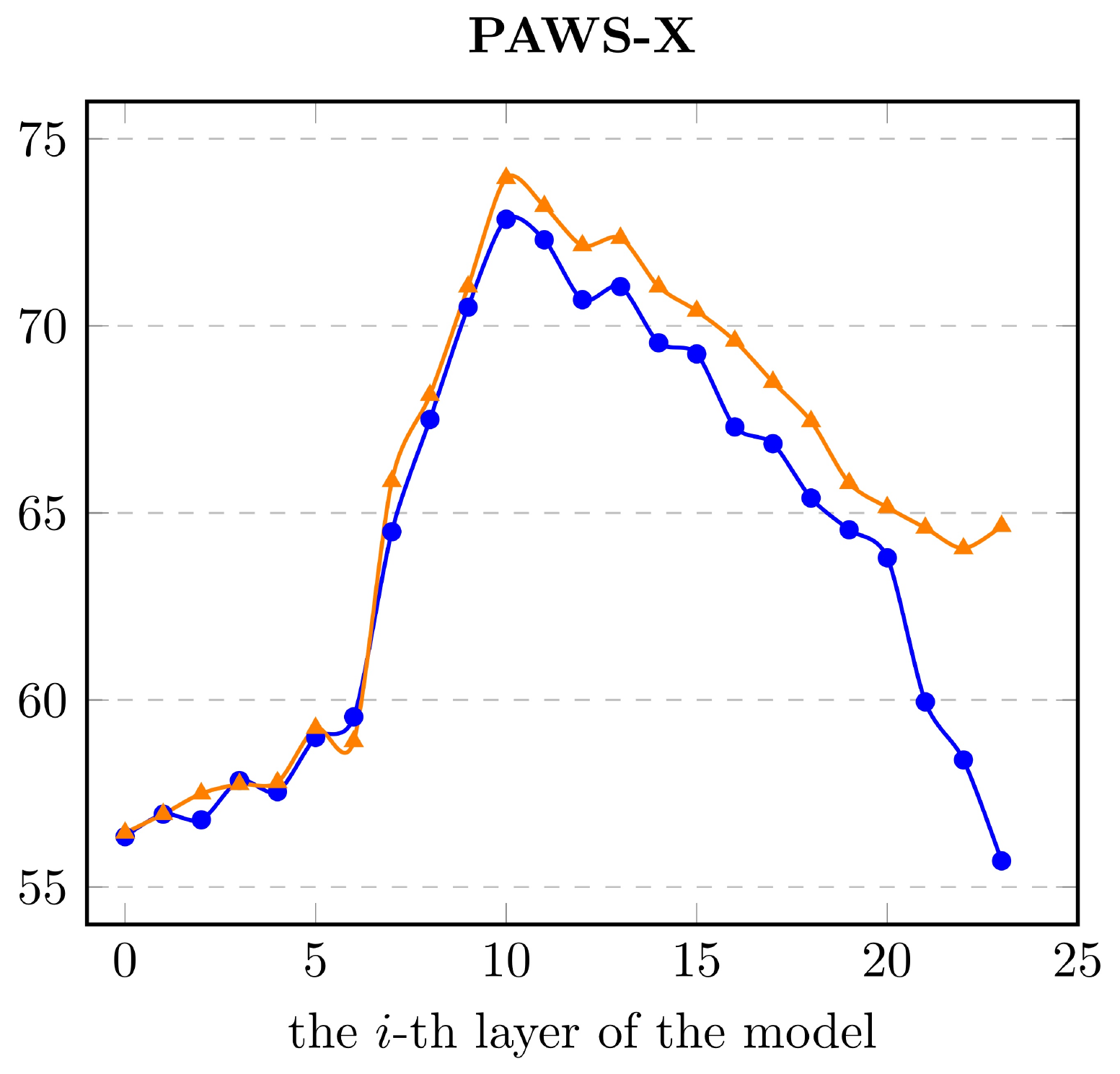}
    \includegraphics[width=0.3\textwidth]{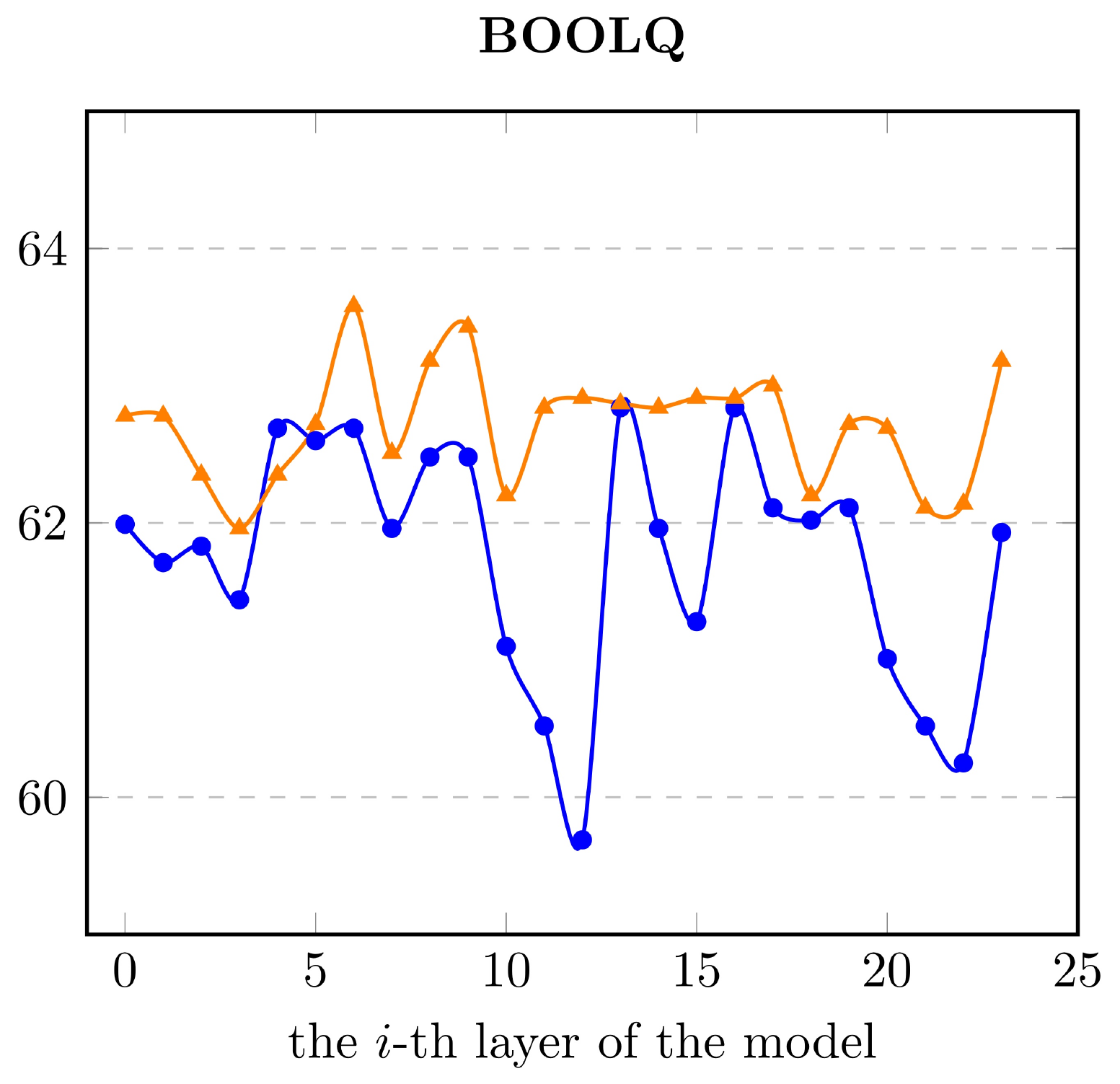}
    \includegraphics[width=0.3\textwidth]{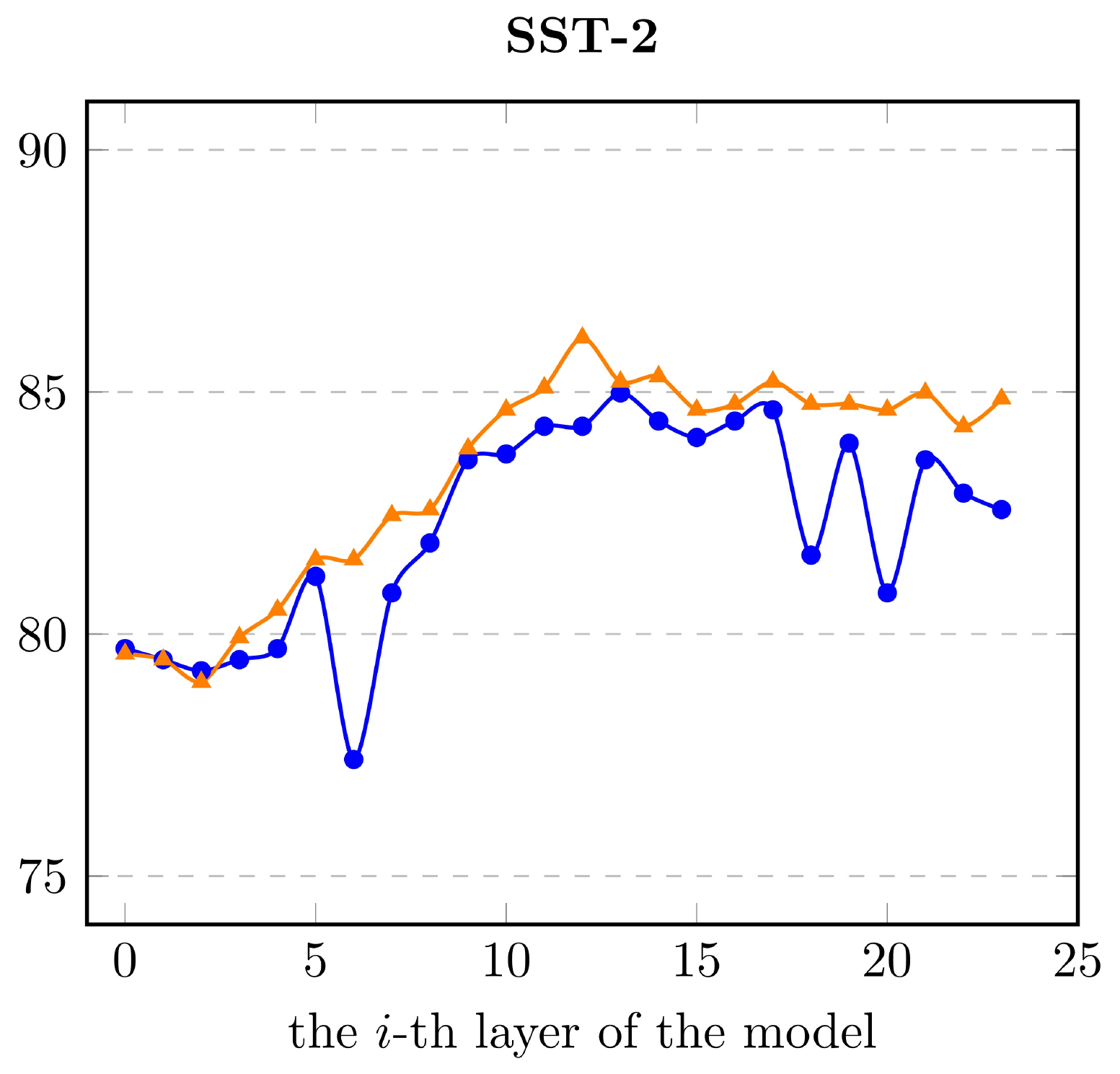}
    
    \includegraphics[width=0.3\textwidth]{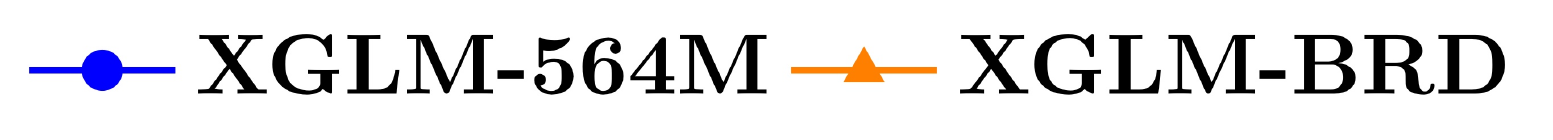}
    
    \caption{The results of probing XGLM-564M and XGLM-BRD layer-wisely on the downstream tasks in the blind test. The horizontal axis represents the specific layer in the model, and the vertical axis is the prediction accuracy (\%) for each task.}
    \label{fig:probing results}
\end{figure*}

Figure \ref{fig:probing results} presents the results of probing XGLM-564M and XGLM-BRD in the blind test. It is clear that XGLM-BRD outperforms XGLM-564M on almost all layers for all downstream tasks. Although XGLM-BRD is trained on the general corpus that is not related to the downstream tasks, basic reading education influences  deep layers of the model, empowering each layer with enhanced downstream task prediction ability.


\subsection{The Impact of Sentiment-related Questions and Answers}

Since our QRA data include questions and answers about the attitude of a sentence, which are related to the SST-2 task, we exclude such data for training XGLM-BRD by deleting the questions about the attitude or the answers containing words of positive/negative/neutral. The objective is to check whether the performance improvement is due to the presence of such data.

Table \ref{tab:without-sentiment-results} shows the result in the blind test. Excluding the sentiment-related data does influence SST-2 performance significantly, resulting in a decrease of 4 points compared to training XGLM-BRD on full data. Thanks to the remaining data for training XGLM-BRD, it still performs significantly better than XGLM-564M by a large margin on SST-2 task. On XNLI task, excluding the sentiment-related data obtains a significant improvement over XGLM-BRD trained on full data. This indicates that the sentiment-related data is not fit for the language inference task.

\begin{table}[t!]
    \centering
    \begin{adjustbox}{width=1\linewidth}
    \begin{tabular}{rccccccc}
    \toprule
    & \multicolumn{6}{c}{Tasks} \\
    \cmidrule{2-7}
    \multicolumn{1}{c}{\textbf{ }} & \textbf{XNLI} & \textbf{RTE}  & \textbf{CB} & \textbf{PAWS-X} & \textbf{BOOLQ} & \textbf{SST-2} & \textbf{Avg}\\
    \midrule
    XGLM-564M   & 35.5 & 46.2 & 53.6 & 51.3 & 51.2 & 63.9 & 50.3 \\
    \midrule
    XGLM-BRD     & 36.2 & 53.8 & \textbf{58.9} & \textbf{56.7} & \textbf{61.0} & \textbf{78.1} & 57.5 \\
    {  $-$SentData} & \textbf{39.2} & \textbf{54.5} & 57.1 & 51.5 & 59.1 & 74.2 & 55.9 \\
    \bottomrule
    \end{tabular}
    \end{adjustbox}
    \caption{The result of training XGLM-BRD based on the data excluding the sentiment-related questions and answers, denoted by $-$SentData, in the blind test.}
    \label{tab:without-sentiment-results}
\end{table}

\end{document}